%% file: main.tex
\documentclass[10pt,twocolumn,letterpaper]{article}

\usepackage[pagenumbers]{iccv} 


\definecolor{iccvblue}{rgb}{0.21,0.49,0.74}
\usepackage[pagebackref,breaklinks,colorlinks,allcolors=iccvblue]{hyperref}

\usepackage{svg}  
\usepackage{graphicx}  
\usepackage{caption}  

\def\paperID{7160} 
\def\confName{ICCV}
\def\confYear{2025}

\title{AIComposer: Any Style and Content Image Composition via Feature Integration}

\author{Haowen Li$^*$\textsuperscript{1,2} \quad Zhenfeng Fan\textsuperscript{2} \quad Zhang Wen\textsuperscript{2} \quad Zhengzhou Zhu\textsuperscript{1} \quad Yunjin Li\textsuperscript{2}\\
\textsuperscript{1}Peking University, China\\
\textsuperscript{2}Beijing Yuanli Science and Technology Co., Ltd., China\\
{\tt\small lhwen@alumni.pku.edu.cn, fanzhenfeng@kanyun.com, wenzhangbj@kanyun.com}\\
{\tt\small zhuzz@pku.edu.cn, liyj@kanyun.com}
}

\begin{document}

\twocolumn[{
    \renewcommand\twocolumn[1][]{#1}
    \maketitle  
    \includegraphics[width=1\textwidth]{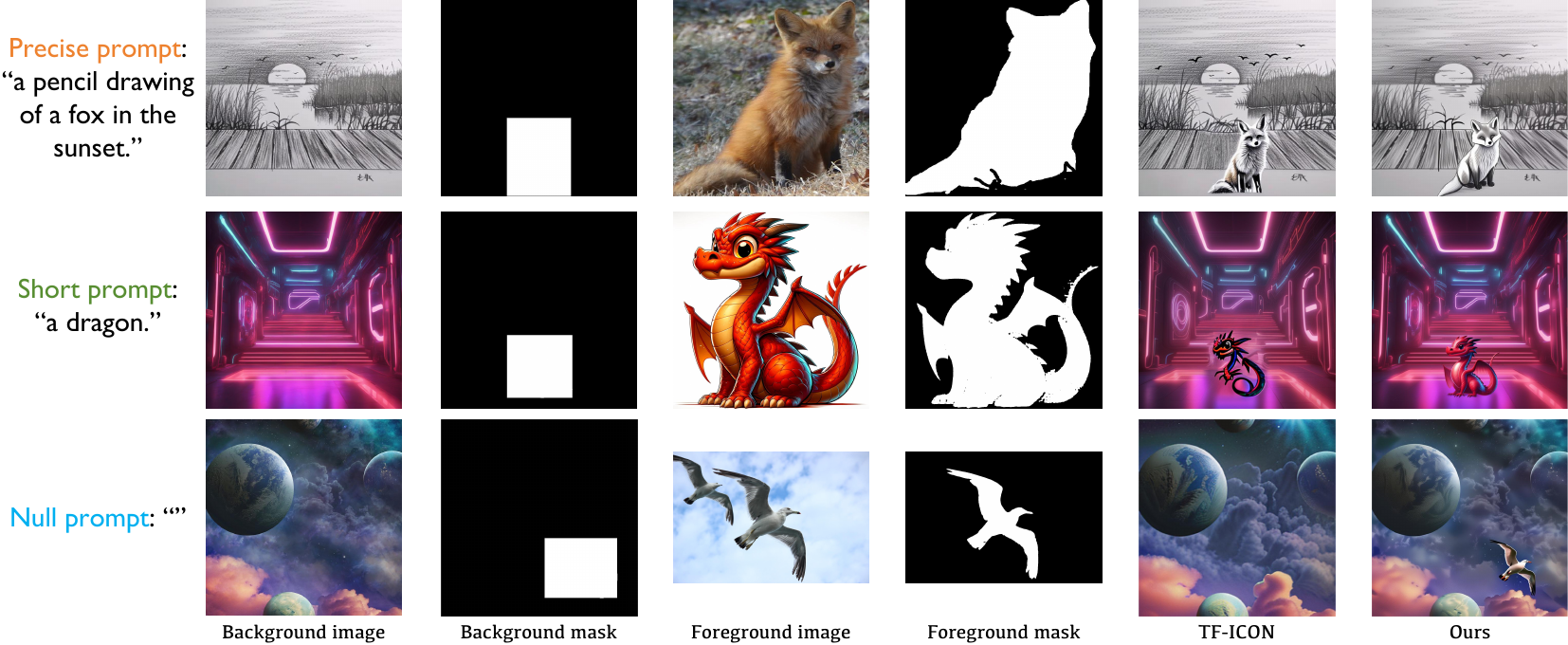} 
    \captionof{figure}{Comparison between our proposed method and a representative state-of-the-art approach TF-ICON \cite{tf-icon} for cross-domain image composition. Both methods deliver good results when the text prompt provides detailed objects and styles. However, our method significantly outperforms TF-ICON in the absence of text prompts describing the style and content. This advancement makes our method highly practical in scenarios where the style and content are difficult to describe with words.}
    \label{fig:first}
    \vspace*{0.3cm}
}]

\renewcommand{\thefootnote}{*} 
\footnotetext{This work was completed during an internship at Beijing Yuanli Science and Technology Co., Ltd.. }
\renewcommand{\thefootnote}{\arabic{footnote}} 

\input{0_abstract}    
\input{1_intro}
\input{2_related-work}
\input{3_approach}

\input{4_experiments}
\input{5_conclusion}

\input{acknowledgment}


 \input{X_suppl}
 {
     \small
     \bibliographystyle{ieeenat_fullname}
     \bibliography{main}
 }

\end{document}

%% file: 0_abstract.tex
\begin{abstract}
Image composition has advanced significantly with large-scale pre-trained T2I diffusion models. Despite progress in same-domain composition, cross-domain composition remains under-explored. The main challenges are the stochastic nature of diffusion models and the style gap between input images, leading to failures and artifacts. Additionally, heavy reliance on text prompts limits practical applications.
This paper presents the first cross-domain image composition method that does not require text prompts, allowing natural stylization and seamless compositions. Our method is efficient and robust, preserving the diffusion prior, as it involves minor steps for backward inversion and forward denoising without training the diffuser. 
Our method also uses a simple multilayer perceptron network to integrate CLIP features from foreground and background, manipulating diffusion with a local cross-attention strategy. It effectively preserves foreground content while enabling stable stylization without a pre-stylization network. Finally, we create a benchmark dataset with diverse contents and styles for fair evaluation, addressing the lack of testing datasets for cross-domain image composition.
Our method outperforms state-of-the-art techniques in both qualitative and quantitative evaluations, significantly improving the LPIPS score by $30.5$\% and the CSD metric by $18.1$\%. We believe our method will advance future research and applications. Code and benchmark at \href{https://github.com/sherlhw/AIComposer}{https://github.com/sherlhw/AIComposer}.
\end{abstract}

%% file: 1_intro.tex
\section{Introduction}
\label{sec:intro}

Image composition, the art of merging a reference (\textit{foreground}) object with a \textit{background}, has greatly progressed with the advent of diffusion models (DM) \cite{DDPM, 2204-06125, stable_diffusion, SahariaCSLWDGLA22}. Some approaches utilize personalized concept learning \cite{dreambooth, CustomDiffusion, textual_inversion, DreamEdit}, attention map manipulations \cite{CustomDiffusion}, and iterative generation \cite{DreamEdit} to synthesize specified images on background images using text prompts. Others incorporate guiding foreground images through self-supervised training to insert a subject while preserving its characteristics by effective data augmentations and content adaptors \cite{objectstitch, paint-by-example}.
Image composition is crucial in entertainment, art, and commerce, however, current methods still exhibit inadequate robustness and fidelity in complicated composited results.

Image composition can be classified into two categories based on style consistency: \textit{same-domain} and \textit{cross-domain} composition. In same-domain composition, the focus is on ensuring smooth edge transitions while retaining the subject's main characteristics, without significant style adjustments. Conversely, cross-domain composition is more challenging, as it requires preserving the subject's core features while adapting its style to harmonize with the background, thereby increasing task complexity.

This paper investigates cross-domain image composition using large-scale diffusion models. A notable challenge in this field is the potential domain gap between images, causing unwanted style inconsistency artifacts, as shown in Figure \ref{fig:qualitative}. 
An intuitive approach is to directly adjust the foreground style to match the background. However, pre-stylizing the foreground before compositing often yields unstable and suboptimal results (Figure~\ref{fig:two-step}). 
This difficulty arises from the confusion of content and style.
Style characteristics are visually perceptible but difficult to precisely describe in language~\cite{InstantStyle}. From this perspective, current approaches~\cite{tf-icon, tale, PrimeComposer} struggle with cross-domain image composition in styles that cannot be articulated through text, as in Figure~\ref{fig:first}.

Given the generative ability of existing large text-to-image (T2I) models with text prompts by Contrastive Language–Image Pre-training (CLIP), we pose the question: \textbf{Can the style as well as the content of images in a composition be managed without a detailed text prompt?} To explore this, we propose integrating the content of the foreground image with the style of the background image through CLIP image features. This way expands the image composition to encompass any content and any style, which is of paramount importance in practical applications. It also mitigates the computational burden and instability often associated with pre-stylizing.

We also aim to minimally modify the original T2I models to preserve the diversity of the diffusion prior. Our method is nearly training-free and entirely tuning-free, meaning that it requires neither additional training for the main diffusion process nor optimization during the test phase. We use a common IP-Adapter~\cite{IP-Adapter} to extract foreground and background features, integrate content and style using a single multilayer perception (MLP) network, and adopt local manipulations step-by-step in the diffusion process. In addition, we improve the inference speed and robustness by utilizing a single-branch structure and fewer diffusion steps. Our approach demonstrates that content and style can be effectively separated and integrated within the CLIP image space, avoiding unreliable explicit pre-stylizing and providing robust outcomes. Furthermore, our method supports scenarios without text prompts when descriptions are unavailable, as in Figure~\ref{fig:first}.

In summary, the main contributions of this paper are:
\begin{itemize}
  \item We propose \textbf{AIComposer}, a cross-domain image composition framework that does not require textual prompts. Our framework seamlessly blends foreground and background images for any content and any style, outperforming existing methods even without precise text prompts.
  \item We adopt a single-branch structure and reduce the inversion and diffusion steps. Our method is simple, efficient, and effective, preserving the diversity of the diffusion prior and leading to robust results compared to existing two-branch, full-step approaches.
  \item We demonstrate that style and content are linearly separable, enabling effective integration within the CLIP features of images. We meticulously combine CLIP features from different images for implicit stylization using a simple MLP network.
  \item We present a publicly available benchmark dataset for cross-domain image composition that features various contents and styles. Our method significantly outperforms existing approaches in both existing and proposed benchmarks.
\end{itemize}

%% file: 2_related-work.tex
\section{Related work}
\label{sec:related-work}
\subsection{Diffusion-based generative models}

Diffusion models \cite{DDPM, 3045358, Song2021, SongE19} have become a leading approach in image generation due to their exceptional capabilities. Modern T2I models~\cite{dreambooth, textual_inversion,BLIP-Diffusion,LayerDiffusion,bld}, primarily employing latent diffusion techniques~\cite{stable_diffusion}, have advanced rapidly in network architectures, high-quality training data, and diverse applications \cite{repaint, imagic, wei2023elite, xiao2024fastcomposer, li2024motrans, shah2025ziplora, brushnet, powerpaint}. These models typically use CLIP features for conditional generation. The original stable diffusion (SD) model~\cite{stable_diffusion} leverages a reduced latent space and cross-attention for conditionality, marking a significant breakthrough in applied research. Within the SD framework, many subsequent works integrate external modules to optimize specific tasks. For instance, Mokady \etal \cite{prompt_to_prompt} manipulate token-wise attention maps for user-controlled image editing. Zhang \etal \cite{controlnet} add a duplicate UNet encoder for conditional control over multiple features. Ye \etal \cite{IP-Adapter} introduce a decoupled cross-attention mechanism, incorporating image prompts alongside text prompts, effectively adapting original images with minor computational overhead.

Our method builds on state-of-the-art T2I models \cite{sdxl, IP-Adapter}, requiring minimal training resources and no tuning during testing. We meticulously design a novel framework for natural stylization and seamless composition that supports any content and style. Additionally, it supports local manipulation, allowing precise control over the diffusion process.

\subsection{Image composition}
Image composition \cite{poisson, abs-2106-14490, controlnet, stable_diffusion, LinYWSL18, abs-2309-15508, objectstitch,  tarresthinking}, \textit{a.k.a} image blending or object inserting, is an enduring challenge in computer vision, with performance rapidly improving thanks to recent large T2I models. 

Some works~\cite{blend_diffusion, bld, SmartBrush, paint-by-example, anydoor, ChenK19} focus on same-domain composition. \eg Yang~\etal~\cite{paint-by-example} train an image-conditioned diffusion model to enable seamless composition on a target image,
Chen~\etal~\cite{anydoor} employ an identity feature extractor
to train a diffusion model using auxiliary videos, 
and Chen \etal \cite{freecompose} 
leverage the original diffusion prior and an optimization step for zero-shot image composition.
Although these methods handle same-domain composition well, they struggle to integrate foreground and background in cross-domain scenarios.

To achieve style consistency in cross-domain image composition, Lu~\etal~\cite{tf-icon} propose a training-free method by manipulating self-attention maps in a two-branch diffusion process. Wang~\etal~\cite{PrimeComposer} refine this by steering self-attention in specific layers and adding local cross-attention for text prompt. Pham~\etal~\cite{tale} enhance style consistency by AdaIN strategies and energy-guided latent optimization in testing. 
In addition, Ruiz~\etal~\cite{magic_insert} proposes a two-stage approach that synthesizes stylized foregrounds before performing adaptive background fusion.
 
A notable advantage of our method is that it breaks the barrier of needing a precise text prompt for content and style. By modifying and injecting CLIP features from images, the method maintains foreground subject characteristics and enables natural stylization without textual descriptions, significantly enhancing image composition for hard-to-describe objects. Additionally, it outperforms existing methods with a more efficient single-branch and fewer-step diffusion process.
 
\subsection{Image style transfer}

The seminal work by Gatys \etal \cite{GatysEB15a, GatysEB15} demonstrates that deep neural networks encode both content and style information of an image. Furthermore, it shows that style and content can be partially separated using an optimization approach involving the Gram matrix. Huang and Belongie \cite{adain} introduce AdaIN, which uses low-order statistics for style transfer. Numerous GAN-based approaches~\cite{cyclegan, AzadiFKWSD18, YangJ0L22, ZhengYZZHZZ24, 0002WRWS24, Zhang0LL17} have also shown effective style transfer results by treating it as a domain adaptation problem. More recently, Wang \etal \cite{InstantStyle} reveal that an image's style is largely tied to specific layers in the diffusion network and propose a style transfer method using pre-trained large diffusion models~\cite{IP-Adapter}.
Building on this, Xing \etal \cite{csgo} customize a dataset with content-style triplets to train large diffusion models for style transfer.

Existing works \cite{Zhang0C18a, ZhangZC20,HongCK21,978-3-031-72684-2_11} demonstrate that style and content are non-linearly separable using specific training objectives and deep neural networks explicitly for images. In this work, we take a step further by showing that the style and content components of an image's CLIP features can be easily and implicitly separated using a linear operation. Based on this, we propose a simple MLP network for style-content integration for the composition task. 

%% file: 3_approach.tex
\section{Methods}
Our objective is to generate a composed image $I^*$ using a background image $I_{bg}$, a foreground image $I_{fg}$ with corresponding binary masks $M_{bg}$ and $M_{fg}$, and an \textbf{optional} text prompt $\mathcal{P}$. 
In this section, we first outline the overall architecture of the proposed framework. Then, we discuss the motivations and novel designs behind the pipeline, including single-branch image inversion and reconstruction, style and content integration, and rectified cross-attention and AdaIN in the diffusion process. 

\subsection{The overall architecture}
\label{sec:3.1}

\begin{figure*}[htbp!]
  \centering
    \includegraphics[width=1\linewidth]{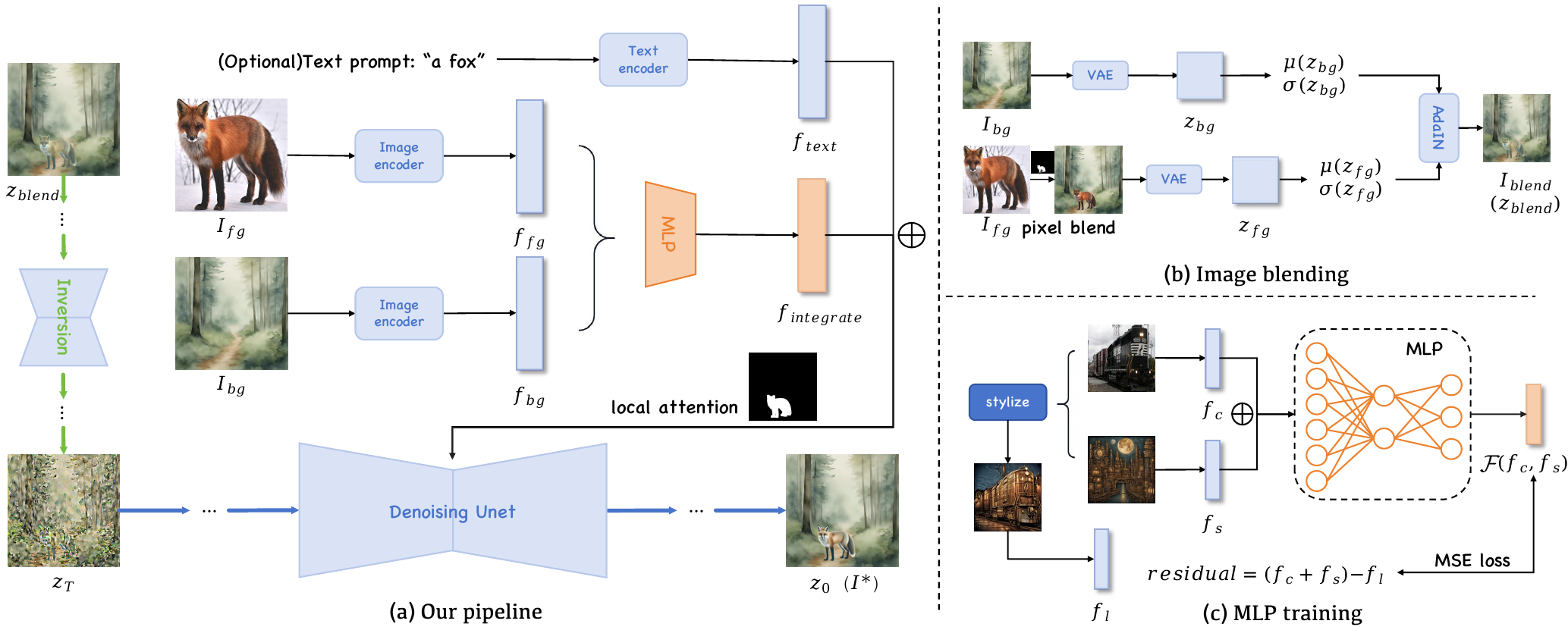}
   \caption{The overall architecture including the (a) pipeline, the (b) image blending strategy, and the (c) MLP training process.}
   \label{fig:pipeline}
\end{figure*}

Stable Diffusion XL (SDXL) \cite{sdxl} is a state-of-the-art large diffusion model, featuring a more sophisticated network architecture and enhanced capabilities compared to its predecessors. As in some existing work \cite{InstantStyle,designedit,csgo}, we utilize the open-source SDXL as our foundational model. 
We also employ the IP-Adapter \cite{IP-Adapter} to extract CLIP image features ($f_{bg}$ and $f_{fg}$) 
, which serve as image prompts to guide the composition process. These image prompts provide \textit{complementary} or \textit{alternative} features to the text prompts. As shown in Figure \ref{fig:pipeline}(b), the foreground $I_{fg}$ and background $I_{bg}$ are initially blended within the latent code of the VAE encoder of SDXL. The blended image $I_{blend}$ (in fact latent $z_{blend}$) is then inverted into a noisy latent representation ($z_T$) using DPM-Solver++~\cite{lu2022dpm}, as in Figure \ref{fig:pipeline}(a). Subsequently, we introduce the integrated image prompt ($f_{integrate}$), derived from both the foreground and background images, along with the text prompt ($f_{text}$), to manipulate cross-attention during specific timesteps of the denoising process. Finally, the denoised latent $z_0$ is decoded by the VAE to generate a composed image $I^*$.

\subsection{Single-branch and fewer-step diffusion}
\label{sec:3.2}

State-of-the-art training-free methods \cite{tf-icon, tale, FreeTuner} typically use a dual-branch diffusion process for image composition, inverting and reconstructing both the foreground and background images to/from noisy latent representations.
Contrary to existing approaches, we find the dual-branch strategy unnecessary. Instead, we conduct a simple and straightforward operation to merge the foreground and background images. Considering that color tone is a substantial part of style, we use AdaIN~\cite{adain} to modulate latent representations, ensuring consistent tones between the foreground and background. 
Please refer to Figure~\ref{fig:pipeline}(b) for the following two steps:

I. Compose the foreground and background images in the pixel space guided by their corresponding masks.

II. Apply an AdaIN operation to the VAE-encoded latent of the foreground ($z_{fg}$) using the background latent ($z_{bg}$) to obtain the initial blended latent ($z_{blend}$).

Figure \ref{fig:qualitative} illustrates some decoded results of the initial blended latent (\ie \textit{InitBlend} in the $6_{th}$ column). The results are marginally acceptable, especially for close-domain compositions. Notably, some examples in Figure \ref{fig:qualitative} reveal that this straightforward strategy can sometimes outperform existing methods that require exquisite operations. 
However, this approach has limitations: it may result in loss of foreground details in some cases and insufficient harmonization of color tones (although with AdaIN) in others. Therefore, it is logical to use this result as a starting point and refine it using additional image and text information.

Our approach is driven by the concept of inverting the blended latent code of the images, rather than inverting the foreground and background images separately.
In this way, we simplify the diffusion process and allow further enhancement using additional content and style information. This method obviates the need to modify the self-attention process or blend the foreground latent \cite{tf-icon, tale, PrimeComposer}. Additionally, we perform inversion and start reconstruction to/from fewer diffusion steps in diffusion, instead of the initial standard Gaussian noise. Consequently, our method simplifies the image reconstruction process, reduces computational overhead, and preserves the original image features as well as the diffusion prior maximally. 

\subsection{Integration of content and style}
\label{sec:3.3}
Wang \etal \cite{InstantStyle} first demonstrate that a large diffusion model can alter the style of an image using a simple IP-Adapter \cite{IP-Adapter}(\ie an adapter for CLIP image features). However, despite carefully injecting the image feature into specific layers of the diffusion model, content preservation is sometimes inadequate. The underlying reason could be that the content and style in the image lack strict definitions and cannot be decoupled thoroughly. 
As a result, directly using the IP-Adapter features of the foreground or background images for image composition is not feasible. We seek a method to separate and combine the CLIP image features of the foreground and background to overcome this difficulty.

Our intuitive idea originates from some characteristics of the CLIP features in T2I diffusion models. When generating a target sample, we typically customize the text prompt with explicit content and style, such as \textit{'a pencil drawing of a panda'}, where \textit{'a panda'} specifies the content and \textit{'a pencil drawing'} specifies the style. This implies that the CLIP text features are well-clustered for content and style.
Then, it is natural to ask \textbf{whether the content and style are also well-clustered within the CLIP image features.} To answer this question, we conducted a simple experiment as follows.

I. Select $80$ images for $20$ styles (contents) from the \textit{Wikiart} dataset \cite{wikiart} (\textit{Caltech-$256$} dataset \cite{Caltech-256}) and use the IP-Adapter to extract their features.

II. Apply Linear Discriminant Analysis (LDA) \cite{lda} and plot the first $2$ dimensions of the resulted features.

\begin{figure}[htbp!]
  \centering
    \includegraphics[width=0.49\linewidth]{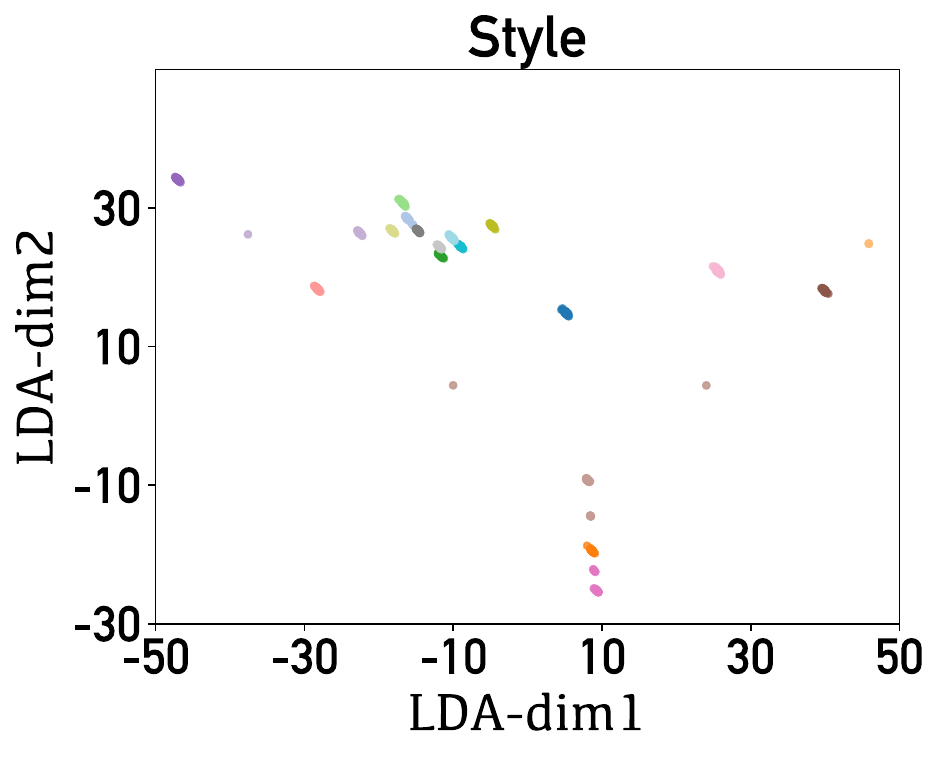}
    \includegraphics[width=0.49\linewidth]{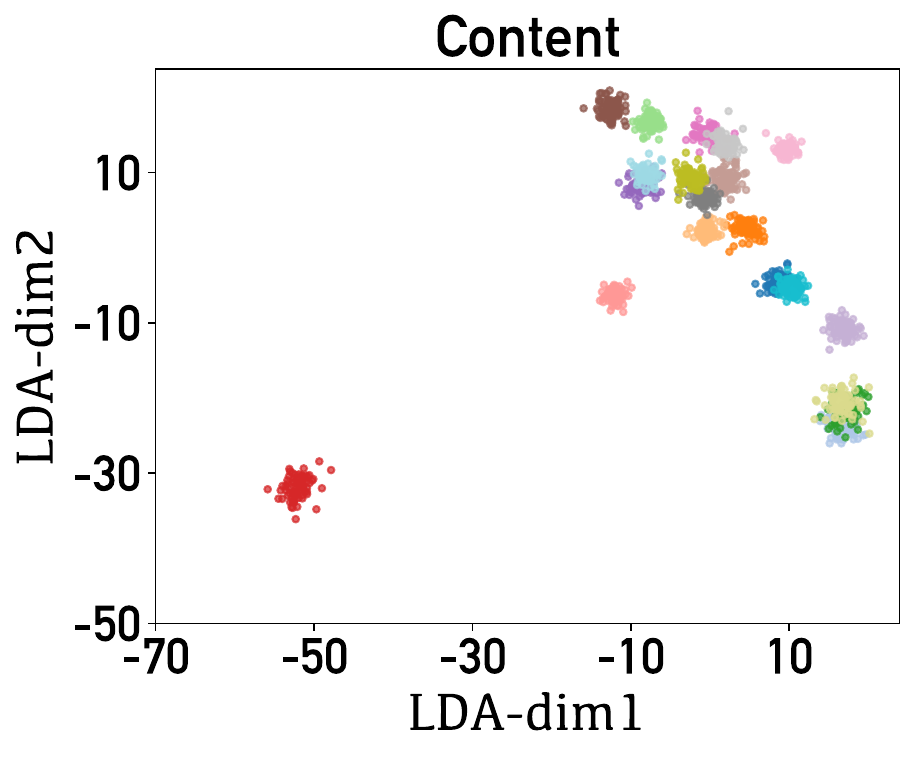}
   \caption{Clustering results of the first two dimensions after applying LDA to the style and content data.
   Different classes are well-separated, albeit not perfectly. The suboptimal result for style, with several clusters within a single class, may indicate that \textit{defining style itself is a challenging task}.}
   \label{fig:lda}
\end{figure}

The dimension-reduced results are illustrated in Figure \ref{fig:lda}. 
It is evident that the content and style features are well-clustered into $20$ classes, even when using the first two dimensions of the simple LDA method. This suggests that merging the content of the foreground with the style of the background within the CLIP image features can be achieved with minor effort.

To enhance this approach, we employ a simple $3$-layer MLP network to elevate the linear operation to a nonlinear one, referring to Figure~\ref{fig:pipeline}(c). 
Training data for the MLP, in the form of (content, style, stylized) triplets, is constructed by an existing work for style transfer \cite{csgo}. 

Let $f_c^k$, $f_s^k$, and $f_l^k$ $(1\leq k \leq N)$ represent the adapted CLIP image features (by IP-Adapter~\cite{IP-Adapter}) for the content, style, and stylized images, respectively. $N$ denotes the number of training samples. The training is to minimize the following objectives:
\begin{equation}\label{e_residual}
  \mathcal{L}=\sum_{1\leq k \leq N}\left\|\mathcal{F}(f_c^k,f_s^k)-(f_c^k+f_s^k-f_l^k) \right\|_2^2,
\end{equation}
where the concatenated features of $f_c^k$ and $f_s^k$ serves as the input of the MLP function $\mathcal{F}(.)$. The supervisory feature $f_c^k+f_s^k-f_l^k$ in Eq.~\ref{e_residual} forces the training in a \textbf{residual} manner. Specifically, we use the feature $f_c^k+f_s^k-\mathcal{F}(f_c^k,f_s^k)$ instead of $\mathcal{F}(f_c^k,f_s^k)$ as $f_{integrate}$ in testing. This strategy leads to better robustness. The reason is that even the $\mathcal{F}(f_c^k,f_s^k)$ is not generalized well (it tends to be random), the network is still able to acquire reliable content and style features\footnote{We visualize the MLP feature in the supplementary material.} from $f_c^k$ and $f_s^k$ considering the additive property of CLIP~\cite{clip}.

With the above efforts, our approach effectively transforms a cross-domain composition problem into a same-domain one. This is similar to a two-step pipeline for performing style transfer before image composition. However, our approach exhibits two primary advantages over a two-step pipeline:

I. We manipulate high-level semantic features instead of explicit original images. These features can be easily decoupled into implicit content and style components, allowing us to use a simple MLP network rather than large style-transfer networks, thereby enhancing efficiency.

\begin{figure}[htbp!]
  \centering
    \includegraphics[width=1\linewidth]{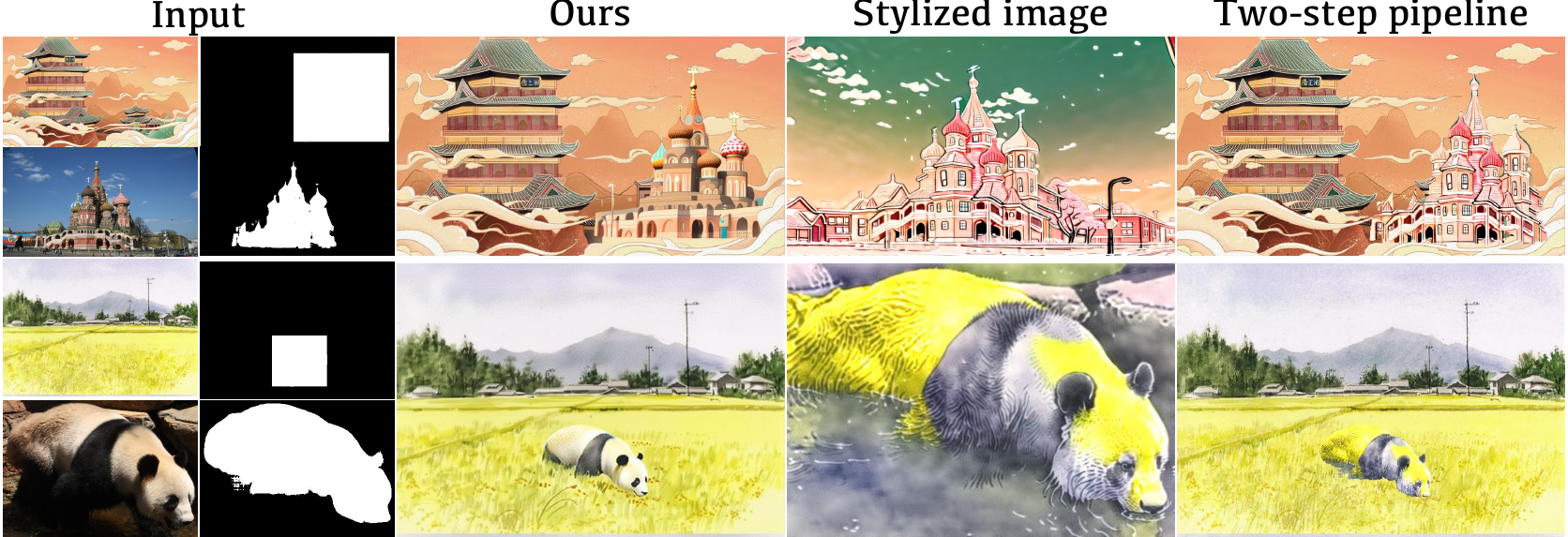}
   \caption{Comparison of our method with the two-step pipeline by applying pre-stylizing \cite{csgo} before same-domain composition.}
   \label{fig:two-step}
\end{figure}

II. Given the difficulty in precisely defining style, even state-of-the-art methods cannot completely avoid poor style-transfer outcomes, as shown in Figure \ref{fig:two-step}. By leveraging the powerful harmonization ability of diffusion models, our approach achieves reliable composition results, even if the stylized features are not perfectly synthesized.
\subsection{Rectified attention control and AdaIN}

In the reconstruction process, we inject the blended feature from the foreground and background as described in Section \ref{sec:3.3}. We follow a recent method~\cite{PrimeComposer} and employ a local strategy to ensure that the injected feature only affects the masked areas. 
To achieve more precise control over the foreground object features, we do not directly use the rectangular mask $M_{bg}$, but instead employ a dilated object mask $M_{dilated}$.
We only modify the areas corresponding to the mask $M_{dilated}$ with additive text features $f_{text}$ and image features $f_{integrate}$ by cross attentions. 

Suppose that $M_{dilated}$ defines a binary function $M$ as
\begin{equation}
  M_{i,j} = 
    \begin{cases}
        0, & i,j \in M_{dilated} \\
        -\infty, & i,j~ \notin ~M_{dilated}
    \end{cases}
\end{equation}
, $Z$ gives the query features in diffusion, and $F$ denotes either $f_{text}$ or $f_{integrate}$. The query, key, and value in cross-attention can be computed by 
$Q = ZW_q, K = FW_k, V = FW_v,$
where $W_q$, $W_k$, and $W_v$ are the weights for modulated features. We omit the subscript of $Z$ as a specific layer and diffusion step for simplicity. The output of the attention operation is rectified by
\begin{equation}\label{attention_e}
  A = Softmax(\frac{QK}{\sqrt{d}}+M)V.
\end{equation}

We denote the output by Eq.~\ref{attention_e} as $A_{integrate}$ and $A_{text}$ for the blended image features and text prompts, respectively. Then, the rectified latent feature in diffusion is
\begin{equation}
  Z_{out} = Z+ A_{integrate} + A_{text}.
\end{equation}

We also adopt the AdaIN operation in diffusion steps to tune the color tone. Let $Z_m$ and $Z_n$ be the masked and unmasked regions of Z, respectively. The AdaIN operation is given by
\begin{equation}
  Z_m \leftarrow (1-\lambda)Z_m + \lambda Z'_m,
\end{equation}
where 
\begin{equation}
  Z'_m=\sigma_n(Z_m-\mu_m)/\sigma_m + \mu_n.
\end{equation}
Here $\mu$ and $\sigma$ denote the standard deviation and the mean, respectively, in which the subscript $m$ and $n$ denote the mask and unmasked regions. The hyperparameter $\lambda$ controls the injecting ratio of AdaIN.

In the reconstruction phase, we only conduct AdaIN and cross-attention operations
in the first $5$ steps. In the later steps, we allow the latent to follow a free diffusion process. We also employ a background preservation strategy by replacing the latent regions outside the mask $M_{dilated}$ with the inverted ones in all steps. This avoids undermining the background due to the stochastic nature of the diffusion process. Overall, we seek an optimal balance among \textit{rigidity}, \textit{manipulability}, and \textit{harmony} with the background of the foreground object.

%% file: 4_experiments.tex
\section{Experiments}

\subsection{Implementation details\protect\footnote{More details are provided in the supplementary material.}}
\label{sec:implementation_details}

\textbf{Training MLP}. As aforementioned in Sec.~\ref{sec:3.3}, our purpose in training MLP is to blend the content information of the foreground and the style information of the background. 
To this end, we selected $401$ images from the ECSSD \cite{ecssd} and $318$ images from the MSRA-B \cite{msra-b} dataset as the content images which represent definite objects. We also select $91$ images of various styles from the Style30k \cite{style30k} and InstantStyle-Plus~\cite{instantstyle-plus}. We construct a triplet for each content and style image using a state-of-the-art work~\cite{csgo} and successfully obtain $65,429$ content-style-stylized triplets. After manually filtering out low-quality and erroneous triplets, we get $37,445$ cleaned data for training. We employ a cross-validation principle to include $N=30,000$ triplets as training data. We utilize the Adam \cite{adam} optimizer with the learning rate of $1 \times 10^{-4}$. The MLP contains only $25.45$M parameters.

\textbf{Other Settings}. We employ DPM-Solver++ \cite{lu2022dpm} inversion with exceptional prompt as in~\cite{tf-icon}. During inference, we resize the background images and masks so that the longest edge is $1024$ pixels to align with the pre-trained SDXL model. The inference process is conducted on an NVIDIA Tesla $32$G-V$100$ GPU. Contrary to existing works \cite{tf-icon, tale, PrimeComposer}, we perform only $10$ steps for inversion and reconstruction, instead of $20$ steps to and from a unit Gaussian noise. 
We set $\lambda = 1$ in the initial image blending stage, and $\lambda = 0.1$ in the diffusion process for AdaIN operations.

\subsection{Testing Benchmarks}
While there are some existing benchmarks~\cite{anydoor, dreambooth, imprint} for same-domain image composition, there is an absence of a benchmark for testing cross-domain image composition. The only one is TF-ICON~\cite{tf-icon}, which includes $95$ examples of cross-domain in all $332$ examples\footnote{The results of same-domain image composition are in the supplementary material.}. Each sample comprises a background image, a foreground image, a background mask, a foreground mask, and a text prompt. We use this as a \textbf{baseline benchmark}, where the background images are in domains (styles) of \textit{pencil sketching}, \textit{oil painting}, and \textit{cartoon animation}. The foreground images are in \textit{photorealism} domain, with limited categories such as animals and food. 

We also extend the above benchmark to include various styles for the background and various contents for the foreground. We select the images from the TF-ICON dataset \cite{tf-icon}, Magic Insert dataset \cite{magic_insert}, and some open-source websites. The foreground masks are obtained with the segment-anything \cite{segany} model. The text prompts of the added data are manually annotated with simple descriptions such as 'a cat', contrary to the TF-ICON baselines (\eg 'a pencil drawing of a panda in the sunset' with definite style and content). As a result, the backgrounds of our \textbf{extended benchmark} include $367$ examples contains various styles such as \textit{photorealistic scenes}, \textit{sketches}, \textit{watercolors}, \textit{oil paintings}, \textit{cartoon animations}, \textit{sci-fi scenes}, \textit{pixel art}, and some others that are not easily defined by language. The resulting foregrounds include $89$ images with various categories such as \textit{animals}, \textit{food}, \textit{cartoon subject}, \textit{buildings}, and some others which are not limited to the photorealism domain. The extended dataset serves as a new cross-domain benchmark for fair and comprehensive evaluations. 

\subsection{Comparison with existing works}

We compare our method with prior SOTA and open-source works capable of performing image composition, including Blended Latent Diffusion (BLD)
\cite{bld}, 
Paint by Example
\cite{paint-by-example}, AnyDoor
\cite{anydoor}, and TF-ICON
\cite{tf-icon}.

\begin{figure*}[htbp!]
  \centering
    \includegraphics[width=1\linewidth]{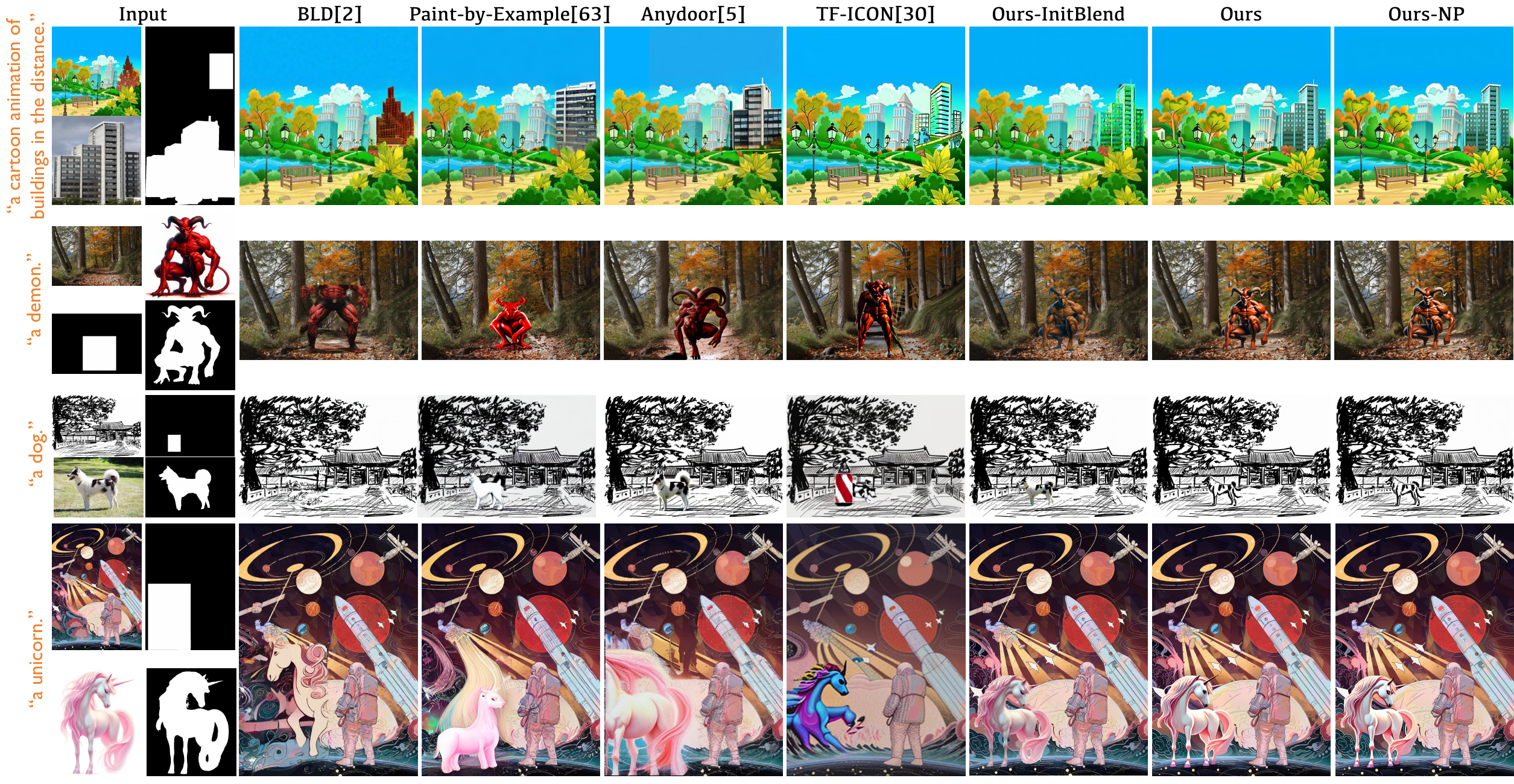}
   \caption{Qualitative comparison of our method with prior SOTA works. We also include the decoded image of the initial blended latent (Ours-InitBlend) and the null-prompt version of our method (Ours-NP) for clearer illustrations.}
   \label{fig:qualitative}
\end{figure*}

\textbf{Evaluation metrics}. Existing research in this field predominantly performs quantitative evaluations of same-domain rather than cross-domain composition, primarily due to the scarcity of established benchmarks. In this work, with proposed testing benchmarks, we are firstly able to perform a fair and comprehensive comparison. To this end, the following metrics are employed for evaluations. 
\begin{itemize}
  \item \textbf{LPIPS}~\cite{lpips} measures the perceptual differences at the patch level between the resulting image and the hybrid image (by directly pasting the foreground image onto the background) within the background masked region $M_{bg}$.
  \item \textbf{CSD}~\cite{csd} measures the style consistency between the resulting image and the background image.
  \item \textbf{PSNR} measures the pixel differences between the resulting image and the background image \emph{outside} $M_{bg}$.
  \item \textbf{CLIP-T}~\cite{clip} similarity measures the alignment score between the resulting image and the given text prompt. 
\end{itemize}
These metrics assess \textit{feature consistency of the painted foreground}, \textit{style consistency between the foreground and background}, \textit{background preservation ability}, and \textit{semantic alignment with the text}, respectively, which are primary goals for cross-domain image composition\footnote{More detailed metrics are in the supplementary material.}.
\begin{table}[htbp!]
	\caption{Quantitative comparisons in the baseline benchmark. The best results are in bold, and the second best are underlined. }
	\vspace{-0.6cm}
	\begin{center}
		\resizebox{0.48\textwidth}{!}{
			\begin{tabular}{lcccc}
                    \toprule
				Method & $\text{LPIPS}_\downarrow$ & $\text{CSD}_\uparrow$ & $\text{PSNR}_\uparrow$ & $\text{CLIP-T}_\uparrow$ \\
				\midrule
				Blended Latent Diffusion \cite{bld} & $0.6536$ & $0.4068$ & {$\underline{21.30}$} & $30.24$ \\
				Paint by Example \cite{paint-by-example} & $0.6167$ & $0.3167$ & $18.04$ & {$\bf{32.85}$} \\
				AnyDoor \cite{anydoor} & $\underline{0.5757}$ & $0.2750$ & $18.63$ & $\underline{32.53}$ \\
                    TF-ICON \cite{tf-icon} & $0.5916$ & $\underline{0.4847}$ & $15.77$ & $32.40$ \\
				Ours & $\bf{0.4281}$ & $\bf{0.4948}$ & $\bf{22.02}$ & $32.12$ \\
                    \hline
                    Ours-NP & $0.4296$ & $0.4743$ & $22.02$ & $31.36$ \\
                    \bottomrule
		\end{tabular}}
	\end{center}
	\vspace{-0.7cm} 
	\label{tab:baseline}
\end{table}
\begin{table}[htbp!]
	\caption{Quantitative comparisons in the extended benchmark. The best results are in bold, and the second best are underlined.}
	\vspace{-0.6cm}
	\begin{center}
		\resizebox{0.48\textwidth}{!}{
			\begin{tabular}{lcccc}
                    \toprule
				Method & $\text{LPIPS}_\downarrow$ & $\text{CSD}_\uparrow$ & $\text{PSNR}_\uparrow$ & $\text{CLIP-T}_\uparrow$ \\
				\midrule
                    Blended Latent Diffusion \cite{bld} & $0.6743$ & $\underline{0.4473}$ & $\underline{18.92}$ & $25.87$ \\
				Paint by Example \cite{paint-by-example} & $0.6684$ & $0.3175$ & $15.14$ & $28.73$ \\
				AnyDoor \cite{anydoor} & $\underline{0.6036}$ & $0.2963$ & $18.25$ & $\underline{29.02}$ \\
                    TF-ICON \cite{tf-icon} & $0.6707$ & $0.4013$ & $13.62$ & $28.23$ \\ 
				Ours & $\bf{0.4195}$ & $\bf{0.5283}$ & $\bf{19.48}$ & $\bf{29.29}$ \\
                    \hline
                    Ours-NP & $0.4221$ & $0.5294$ & $19.47$ & $28.75$ \\
				\bottomrule
		\end{tabular}}
	\end{center}
	\vspace{-0.7cm} 
	\label{tab:extended}
\end{table}

Table~\ref{tab:baseline} and Table~\ref{tab:extended} show the quantitative results of the baseline benchmark and the extended benchmark, respectively. We also include a version of our method without text prompts, \ie \textit{Ours-NP}. As observed, our method achieves significantly better performance over the existing works even without text prompts. In the baseline benchmark, our method achieves the best results in terms of all metrics except for CLIP-T. Paint by Example~\cite{paint-by-example} and AnyDoor~\cite{anydoor} lead to better CLIP-T because they are closer to a copy-paste method without considering style consistency. We argue that each metric alone may not reveal its effectiveness.
In the extended benchmark, our method significantly surpasses existing work in all metrics. Figure~\ref{fig:qualitative} shows some typical examples. We can observe that our method
1) naturally stylizes the subject according to the background image without relying on text prompts; 2) better preserves the identity characteristics of subjects that are hard to describe with language; and 3) enables seamless integration of objects into the background without noticeable artifacts in the transition areas. In conclusion, we achieve much more robust results than existing works, even without text prompts and with a simpler diffusion process. 
We owe the success to our single-branch and fewer-step strategy against undermining the image features and diffusion priors, and the MLP feature implicit blending method for effective stylization. 

\subsection{User study}
One of our findings in studying the cross-domain image composition problem is that adapting the foreground color before pasting it onto the background yields good quantitative results, despite unsatisfactory qualitative outcomes. Therefore, the quantitative metrics may have 
\begin{figure}[htbp!]
  \centering
    \includegraphics[width=1\linewidth]{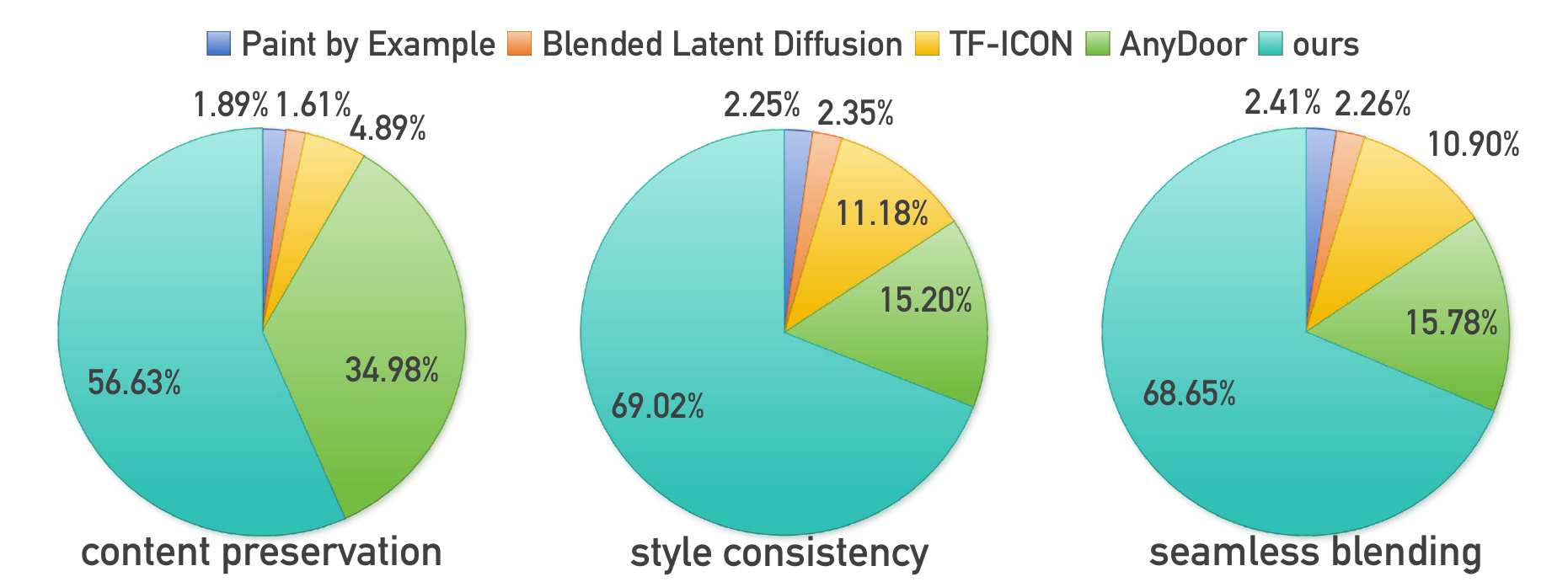}
   \caption{The average user preference percentage for different methods in the extended benchmark.}
   \label{fig:user_study}
\end{figure}
their limitations. To assess the results more comprehensively, we conduct a user study by asking $50$ participants to select the best results in \textit{content preservation}, \textit{style consistency}, and \textit{seamless blending} among different methods in the extended benchmark. Figure~\ref{fig:user_study} shows that our method is favored by most participants on average, indicating its superiority.

\subsection{Ablation study and additional analysis}

The elements of our pipeline on the extended dataset with the following configurations: ($1$) our method without CLIP image features (-ImageCLIP); ($2$) ours without initial latent blending (-InitBlend) (we blend the images in pixels instead); ($3$) ours without inversion (-Inversion); ($4$) ours with full diffusion steps (FullDiffusion); ($5$) ours with a dual-branch inversion and diffusion strategy (DualBranch). The quantitative results are shown in Table \ref{tab:efficiency}. 
\begin{figure}[htbp!]
  \centering
    \includegraphics[width=1\linewidth]{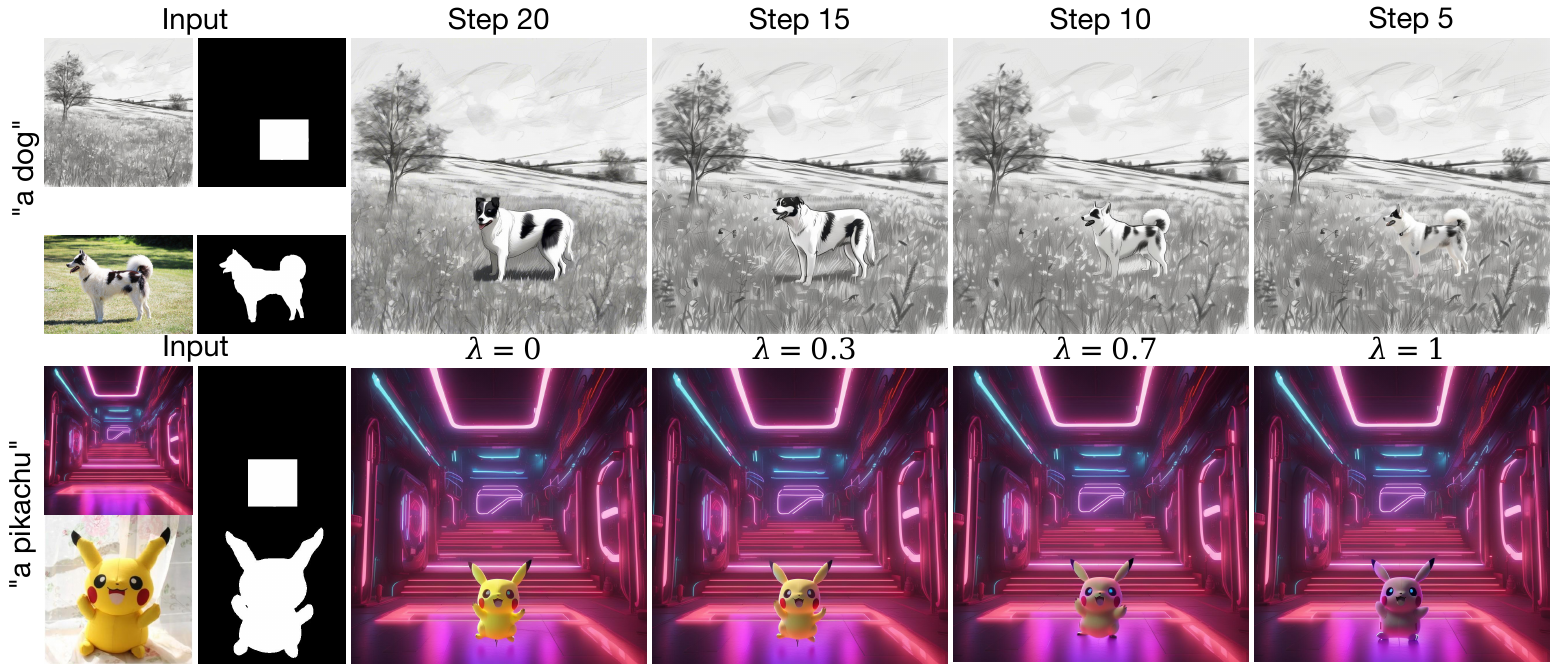}
   \caption{Parameter Impact Analysis: (Top) different diffusion steps; (Bottom) different AdaIN radio $\lambda$ for initial blending.}
   \label{fig:parameter}
\end{figure}
As observed, Inversion preserves foreground subject characteristics, ImageCLIP leads to noticeable improvements in LPIPS, CSD, and CLIP-T, while InitBlend significantly enhances the CSD metric. Our method without InitBlend achieves better LPIPS and CLIP scores, indicating the limitations of these metrics as previously mentioned. 
Table \ref{tab:efficiency} demonstrates that a FullDiffusion or a DualBranch strategy is unnecessary since they introduce additional computational overhead without improving performance. The ImageCLIP and InitBlend strategies in our method require minor computational costs as in Table~\ref{tab:efficiency}. The above experiments demonstrate that our method is simple, efficient, yet effective in its designs.
In addition, Figure \ref{fig:parameter} shows adjusting the diffusion steps could better balance rigidity and variations (\eg shadows and detailed variations), while the AdaIN ratio $\lambda$ for initial blending can control the intensity of color stylization.

\begin{table}[htbp!]
	\caption{Ablation study on the effectiveness of each component. The best results are in bold, and the second best are underlined.}
	\vspace{-0.6cm}
	\begin{center}
		\resizebox{0.4\textwidth}{!}{
			\begin{tabular}{lcccc}
				\toprule
				Method & $\text{LPIPS}_\downarrow$ & $\text{CSD}_\uparrow$ & $\text{CLIP-T}_\uparrow$ & $\text{Time}_\downarrow$ \\
				\midrule
				Baseline & $\underline{0.4195}$ & $\bf{0.5284}$ & $\underline{28.75}$ & $12.94$s \\
				-ImageCLIP & $0.4208$ & $0.5083$ & $28.66$ & \textbf{$\bf11.39$s} \\
				-InitBlend & $\bf{0.3376}$ & $0.4671$ & $\bf{28.98}$ & \underline{$12.92$s} \\  
                    -Inversion & $0.5867$ & $0.4509$ & $24.23$ & $13.21$s \\  
                    FullDiffusion  & $0.4763$ & $0.5053$ & $28.51$ & $19.67$s \\
                    DualBranch  & $0.4945$ & \underline{$0.5219$} & $26.49$ & $29.25$s \\       
				\bottomrule
		\end{tabular}}
	\end{center}
	\vspace{-0.7cm} 
	\label{tab:efficiency}
\end{table}


%% file: 5_conclusion.tex
\section{Limitations}
Although we achieve robust results for cross-domain image composition without a text prompt, our approach has limitations. Since our approach relies on pre-trained large diffusion models without diffuser training or test-time optimization, it inherits their shortcomings and biases. Real-time performance is difficult to achieve, and the results may be inconsistent due to the stochastic nature of diffusion.
As diffusion models continue to evolve on both large and small scales, we anticipate further performance improvements. 

\section{Conclusion}
In this paper, we introduce a novel framework for cross-domain image composition that does not rely on text prompts, effectively addressing the limitations of current methods. By using a MLP network to integrate content and style within the CLIP image space, we achieve image composition with high-fidelity content, natural stylization, and seamless blending. Our framework does not require diffuser training or test-time optimization, but employs a single-branch and fewer-step structure in diffusion, significantly improving the robustness and efficiency.
Generally, the proposed framework is highly practical in scenarios where style and content are difficult to describe using language.

%% file: acknowledgment.tex
\section{Acknowledgment}
This paper is supported by the Humanities and Social Sciences Research Planning Fund Project of the Ministry of Education: "Research on Metacognitive Diagnosis Theory and Technology Driven by Multimodal Learning Data" (23YJA880091).

%% file: X_suppl.tex
\clearpage
\setcounter{page}{1}
\maketitlesupplementary

\section{Ablations and Additional analysis}
\subsection{Qualitative analysis of ablations}
\begin{figure*}[htbp!]
  \centering
    \includegraphics[width=0.8\linewidth]{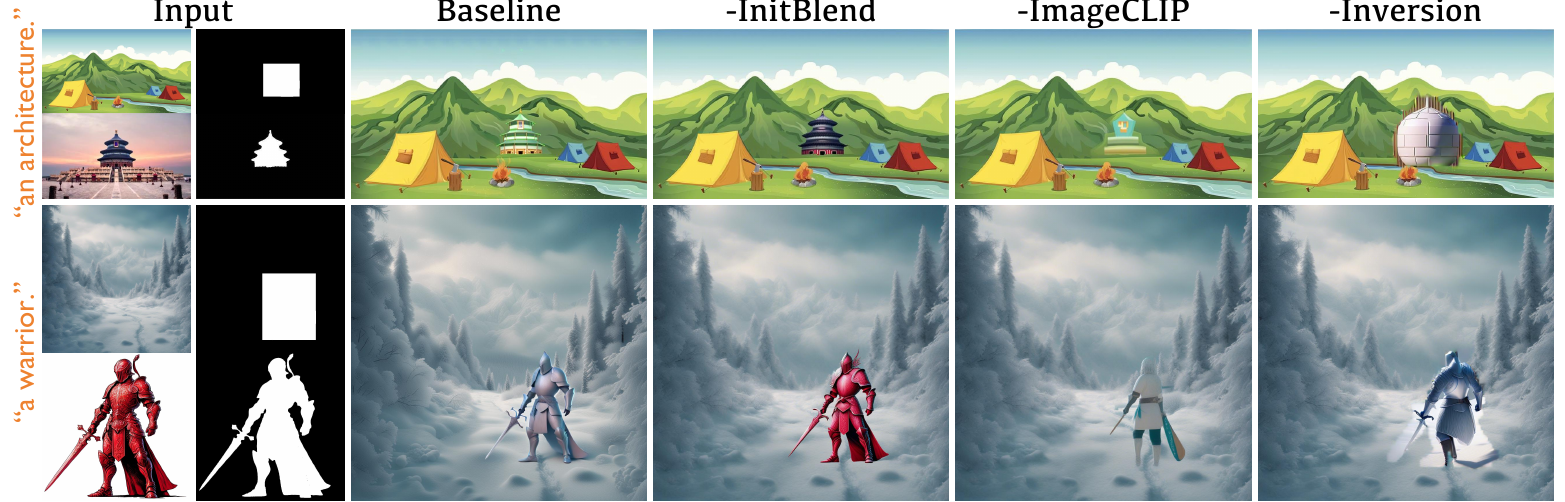}
   \caption{The qualitative results of ablation study.}
   \label{fig:ablation_qualitative}
\end{figure*}

Figure~\ref{fig:ablation_qualitative} presents qualitative comparisons of our ablation study. The removal of inversion (-Inversion) exhibits noticeable degradation in foreground subject content features and introduces incoherence between the foreground and background. Similarly, the exclusion of ImageCLIP (-ImageCLIP) similarly demonstrates a substantial impact on the characteristics of the foreground subject. Additionally, -InitBlend significantly compromises the style transfer quality of foreground subjects.

\subsection{Visualizing the features of MLP}
\begin{figure}[htbp!]
  \centering
    \includegraphics[width=1\linewidth]{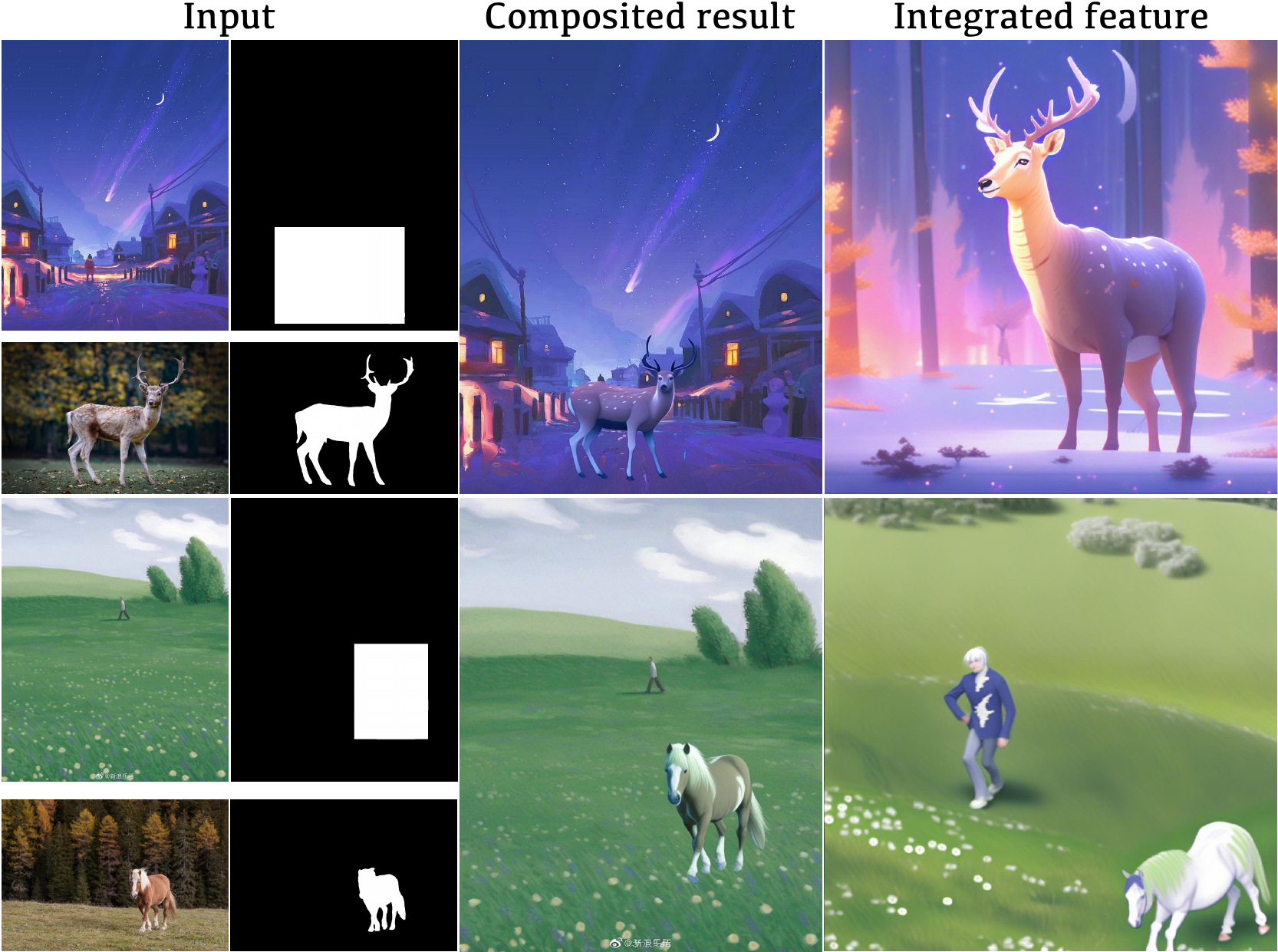}
   \caption{Two examples of the composited results and their corresponding integrated features by MLP (visualizing with IP-Adapter~\cite{IP-Adapter}). The top example is well pre-stylized, whereas the bottom example shows content leakage of a man in the integrated feature.}
   \label{fig:show_mlp1}
\end{figure}

In this work, we integrate the features of foreground content and background style with a simple MLP network. Therefore, the MLP network implicitly serves for style transfer. Although we can not visualize the pre-stylized image directly, the integrated features by MLP can be visualized with an IP-Adapter~\cite{IP-Adapter}. Figure~\ref{fig:show_mlp1} presents two typical examples of the composited results and the integrated features. We anticipate that the integrated feature captures the content of the foreground while pre-stylizing it referring to the 
background. In most cases, such as the top example in Figure~\ref{fig:show_mlp1}, the MLP network effectively performs implicit pre-stylizing. In other cases, such as the bottom example in Figure~\ref{fig:show_mlp1}, even if the pre-stylizing features are not perfectly blended, our method still achieves robust results. This demonstrates the effectiveness of our feature blending strategy.

\subsection{Effects of residual learning and number of triplet data}

In this work, we employ $30,000$ triplets for training MLP in a \textbf{residual} manner, ensuring robust learning for various contents and styles.
To learn the effect of the number of training data and the residual training strategy, we conduct additional experiments, including $300$ and $3,000$ training data. We also evaluated the effect of the residual learning strategy by replacing it with direct learning for blended features.
\begin{figure*}[htbp!]
  \centering
    \includegraphics[width=1\linewidth]{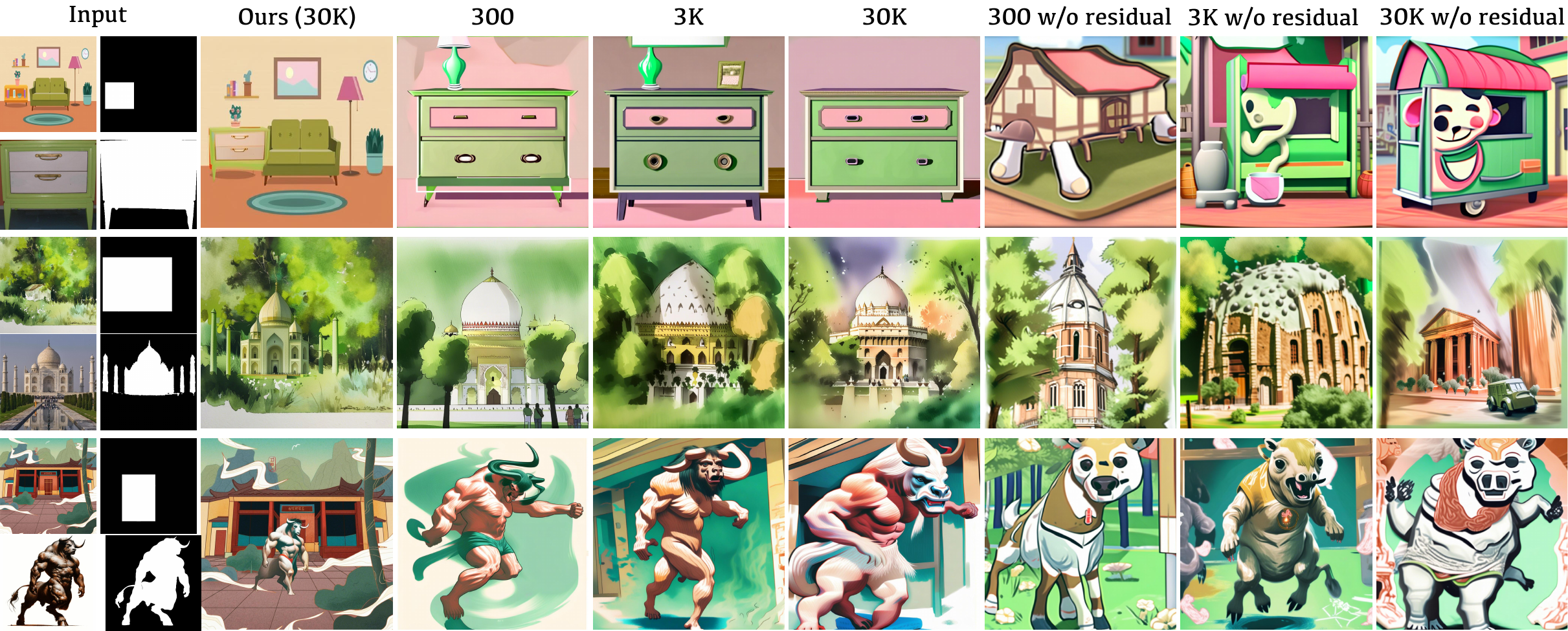}
   \caption{Examples of our composited results (the $2_{nd}$ column), and blended feature visualization (the last $6$ columns) using different number of training data, with and without residual training strategy.}
   \label{fig:show_mlp}
\end{figure*}
Figure~\ref{fig:show_mlp} shows some examples of these settings. We visualize the integrated features by an IP-Adapter~\cite{IP-Adapter}. It is evident that the residual learning strategy contributes to robustness in the integration of the features of foreground content and background style. Figure~\ref{fig:show_mlp} also indicates that the strong generalization performance of the feature blending strategy does not depend on a large number of training triplets. We attribute this success to our meticulous design of residual learning, which takes full consideration of the inherent additive properties of CLIP~\cite{clip}. During inference, we add the CLIP features of the foreground and background images, and then subtract a learned residual. Even if the residual is not well learned (tending towards random features), our method can still effectively capture most of the content and style features from the input foreground and background images. This enhances the robustness of our method.

To validate the impact of different stylization methods on MLP performance, we additionally generated a subset of training data using StyleID~\cite{ChungHH24}, with corresponding quantitative results presented in Table~\ref{tab:styleid}. The results indicate that different stylization models have no significant impact on our method's performance, while data filtering and increased dataset size consistently improve results. This demonstrates the robustness of our approach across stylistic variations.

\begin{table}[htbp!]
	\caption{Ablation study on the various data sources and stylization models. The best results are in bold, and the second best are underlined.}
	\vspace{-0.6cm}
	\begin{center}
		\resizebox{0.48\textwidth}{!}{
			\begin{tabular}{lccccc}
				\toprule
                Data Num & Source & Filtered & $\text{LPIPS}_\downarrow$ & $\text{CSD}_\uparrow$ & $\text{CLIP-T}_\uparrow$  \\ 
                    \midrule
                    30K & CSGO~\cite{csgo} & Yes & $\bf{0.4195}$ & $\bf{0.5284}$ & $\bf{28.75}$ \\
				300 & CSGO~\cite{csgo} & Yes & $0.4322$ & $\underline{0.5227}$ & $28.05$  \\
                    300 & CSGO~\cite{csgo} & No & ${0.4354}$ & $0.5159$ & ${28.01}$ \\
                    30K & StyleID~\cite{ChungHH24} & No & $0.4340$ & $0.5154$ & $28.07$\\
                    300 & StyleID~\cite{ChungHH24} & Yes & $\underline{0.4291}$ & $0.5152$ & $\underline{28.21}$ \\
                    300 & StyleID~\cite{ChungHH24} & No & $0.4327$ & $0.5133$ & $28.15$ \\
                    
				\bottomrule
		\end{tabular}}
	\end{center}
	\vspace{-0.7cm} 
	\label{tab:styleid}
\end{table}


\section{Results for same-domain composition}
Despite its powerful capabilities in handling cross-domain image composition, our method can also be applied to same-domain cases. Table~\ref{tab:same-domain} summarizes the quantitative results on the same-domain benchmark of TF-ICON~\cite{tf-icon}, as mentioned in the body text of this paper. We include the common CLIP similarity metrics for text prompts and foreground images, as well as LPIPS scores~\cite{lpips} for both foreground and background images. Our methods achieve significantly better LPIPS scores for the composed foreground, indicating superior preservation of foreground content over existing works. Some other metrics for our method are also competitive with the existing works. This demonstrates that our approach is robust in handling same-domain image compositions as well as the cross-domain cases detailed in the main text of this paper.

\begin{table}[htbp!]
	\caption{Quantitative comparisons in the remaining $237$ samples for the same-domain (photorealism) TF-ICON benchmark. The best results are in bold, and the second best are underlined.}
	\vspace{-0.6cm}
	\begin{center}
		\resizebox{0.48\textwidth}{!}{
			\begin{tabular}{lcccc}
                    \toprule
				Method & $\text{CLIP}_\text{(Text)}\uparrow$ & $\text{CLIP}_\text{(Image)}\uparrow$ & $\text{LPIPS}_\text{(BG)}\downarrow$ & $\text{LPIPS}_\text{(FG)}\downarrow$  \\
				\midrule
				Blended \cite{bld} & $25.19$ & $73.25$ & $0.11$ & $0.77$ \\
				Pa. by Ex. \cite{paint-by-example} & $25.92$ & $80.26$ & $0.13$ & $0.73$ \\
				AnyDoor \cite{anydoor} & $\bf{31.24}$ & $\bf{87.87}$ & $\underline{0.09}$ & $0.59$ \\
                    TF-ICON \cite{tf-icon} & $28.11$ & $82.86$ & $0.10$ & $0.60$  \\
                    TALE \cite{tale} & $31.03$ & $\underline{85.12}$ & $0.10$ & $0.51$  \\
                    PrimeCom. \cite{PrimeComposer} & $30.26$ & $84.71$ & $\bf{0.08}$ & $\underline{0.48}$  \\
				Ours & $\underline{31.08}$ & $84.26$ & $\bf{0.08}$ & $\bf{0.36}$ \\
                    \bottomrule
		\end{tabular}}
	\end{center}
	\vspace{-0.7cm} 
	\label{tab:same-domain}
\end{table}

\section{More results for cross-domain composition}
We include $3$ additional metrics to assess the results of different methods in our extended benchmark. 
\begin{itemize}
  \item \textbf{FID}~\cite{fid} and \textbf{ArtFID}~\cite{artfid} measure the content and style similarity between the resulting image and the hybrid image (by directly pasting the foreground image onto the background image) within the background masked region $M_{bg}$.
  \item \textbf{CLIP-I}~\cite{clip} similarity measures the alignment score between the background masked region of the resulting image and the foreground image. 
\end{itemize}
The FID and ArtFID metrics are also effective in measuring the composited results in both content preservation and style consistency. CLIP-I acts similarly to CLIP-T, which is commonly used in same-domain image composition. The quantitative results for different methods are included in Table~\ref{tab:all-metrics}.
We can see that our method suppresses the existing ones except for the CLIP-I score. We argue that the CLIP-I metric focuses more on content preservation and does not account for style transformation.

We also implement our method with an early version of stable diffusion (\textit{ours-SD1.5})~\cite{stable_diffusion} for more comprehensive evaluations. 
Although an advanced SDXL model~\cite{sdxl} leads to some gains in quantitative performance, combining our method with an early version of SD also outperforms the existing works by a significant margin. 
This demonstrates that the enhanced performance of our method is primarily due to MLP feature blending and single-branch diffusion strategies.

\begin{figure*}[htbp!]
  \centering
    \includegraphics[width=1\linewidth]{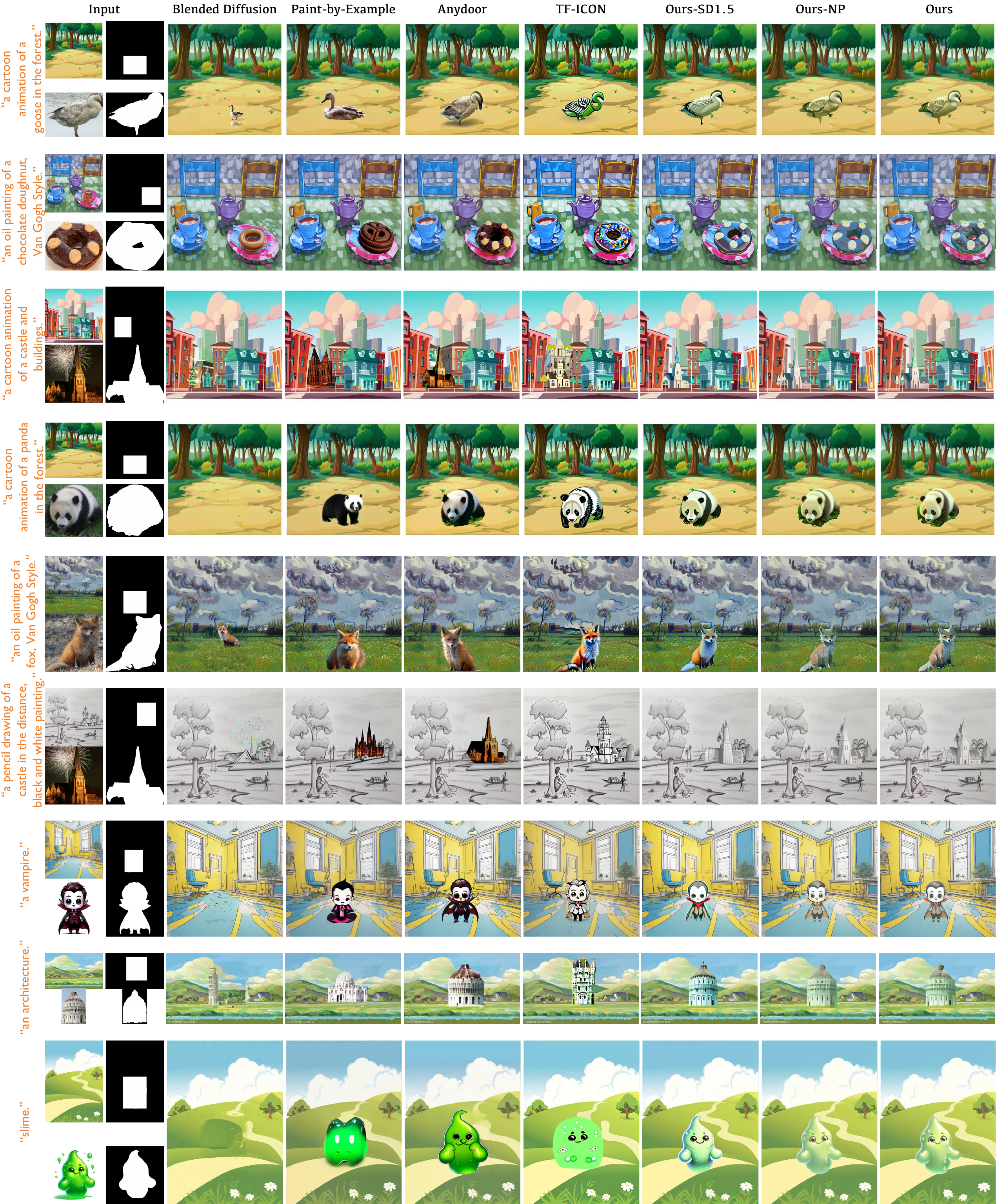}
   \caption{Qualitative comparison of our method with prior SOTA works. We also include an early version of stable diffusion (Ours-SD$1.5$) and the null-prompt version for our method (Ours-NP).}
   \label{fig:1}
\end{figure*}

\begin{figure*}[htbp!]
  \centering
    \includegraphics[width=1\linewidth]{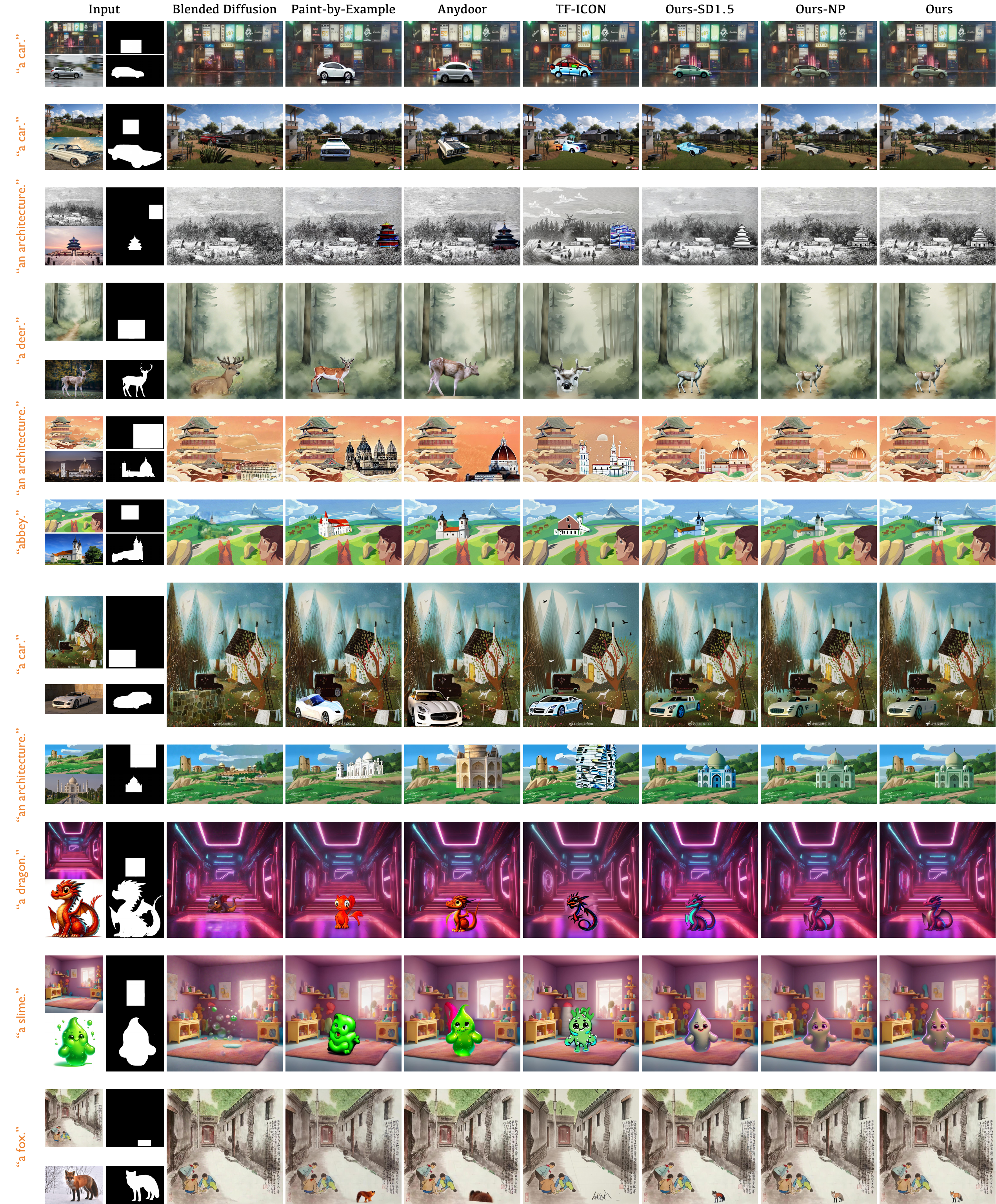}
   \caption{Qualitative comparison of our method with prior SOTA works. We also include an early version of stable diffusion (Ours-SD$1.5$) and the null-prompt version for our method (Ours-NP).}
   \label{fig:2}
\end{figure*}

\begin{figure*}[htbp!]
  \centering
    \includegraphics[width=1\linewidth]{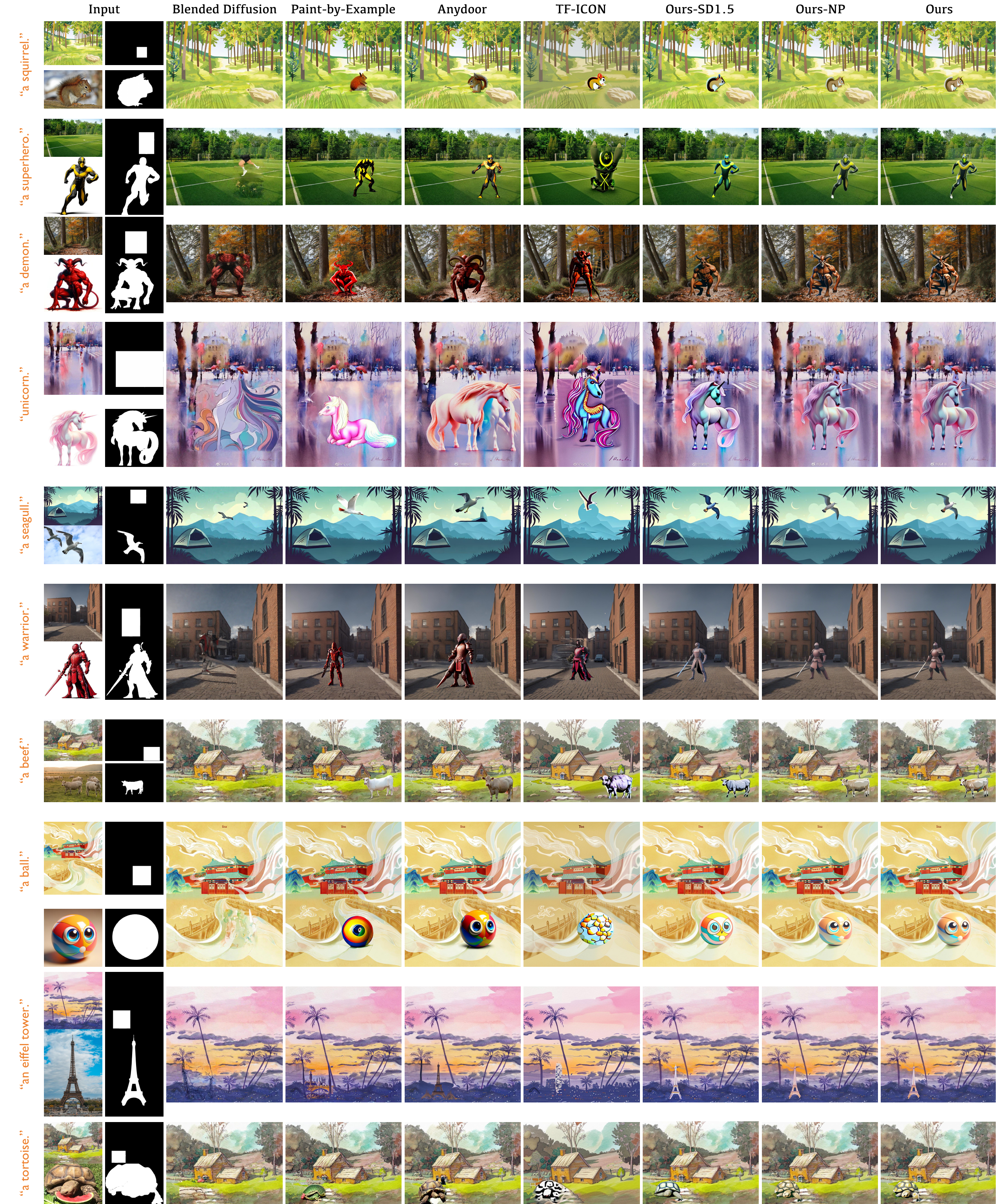}
   \caption{Qualitative comparison of our method with prior SOTA works. We also include an early version of stable diffusion (Ours-SD$1.5$) and the null-prompt version for our method (Ours-NP).}
   \label{fig:3}
\end{figure*}

More qualitative results are in Figure~\ref{fig:1}, Figure~\ref{fig:2}, and Figure~\ref{fig:3}. With these versatile examples, we hope that readers can gain a more intuitive understanding of our methods and our proposed benchmark dataset for cross-domain image composition.

\begin{table*}[htbp!]
	\caption{Quantitative comparisons in the extended benchmark. The best results are in bold, and the second best are underlined. We also include an early version of stable diffusion (Ours-SD$1.5$) and the null-prompt version for our method (Ours-NP).}
	\vspace{-0.6cm}
	\begin{center}
		\resizebox{0.7\textwidth}{!}{
			\begin{tabular}{lcccccccc}
                    \toprule
				Method & $\text{BaseModel}$ & $\text{LPIPS}_\downarrow$ & $\text{FID}_\downarrow$ & $\text{ArtFID}_\downarrow$ & $\text{CSD}_\uparrow$ & $\text{PSNR}_\uparrow$ & $\text{CLIP-T}_\uparrow$ & $\text{CLIP-I}_\uparrow$  \\
				\midrule
				Blended \cite{bld} & SDXL & $0.6743$ & $13.77$ & $24.74$ & $\underline{0.4473}$ & $\underline{18.92}$ & $25.87$ & $66.33$ \\
				Paint by Example \cite{paint-by-example} & SD$1.4$ & $0.6684$ & $12.29$ & $22.17$ & $0.3175$ & $15.14$  & $28.73$ & $\underline{78.94}$ \\
				AnyDoor \cite{anydoor} & SD$2.1$ & $\underline{0.6036}$ & $\underline{9.587}$ & $\underline{16.95}$ & ${0.2963}$ & $18.24$ & $\underline{29.02}$ & $\bf{84.40}$   \\
                    TF-ICON \cite{tf-icon} & SD$2.1$ & $0.6707$ & $12.82$ & $23.05$ & $0.4013$ & $13.62$ & $28.23$ & $70.98$   \\
				Ours & SDXL & $\bf{0.4195}$ & $\bf{9.538}$ & $\bf{14.97}$ & $\bf{0.5283}$ & $\bf{19.48}$ & $\bf{29.29}$ & ${77.59}$  \\
                    Ours-NP & SDXL & $0.4221$ & $9.682$ & $15.20$ & $0.5294$ & $19.49$ & $28.75$ & $75.93$  \\
                    Ours-SD1.5 & SD$1.5$ & $0.5093$ & $11.15$ & $18.31$ & $0.4886$ & $16.63$ & $28.79$ & $74.40$   \\
                    \bottomrule
		\end{tabular}}
	\end{center}
	\vspace{-0.7cm} 
	\label{tab:all-metrics}
\end{table*}

\section{Extensions}

\subsection{Combination with ControlNet}
\begin{figure}[htbp!]
  \centering
    \includegraphics[width=1\linewidth]{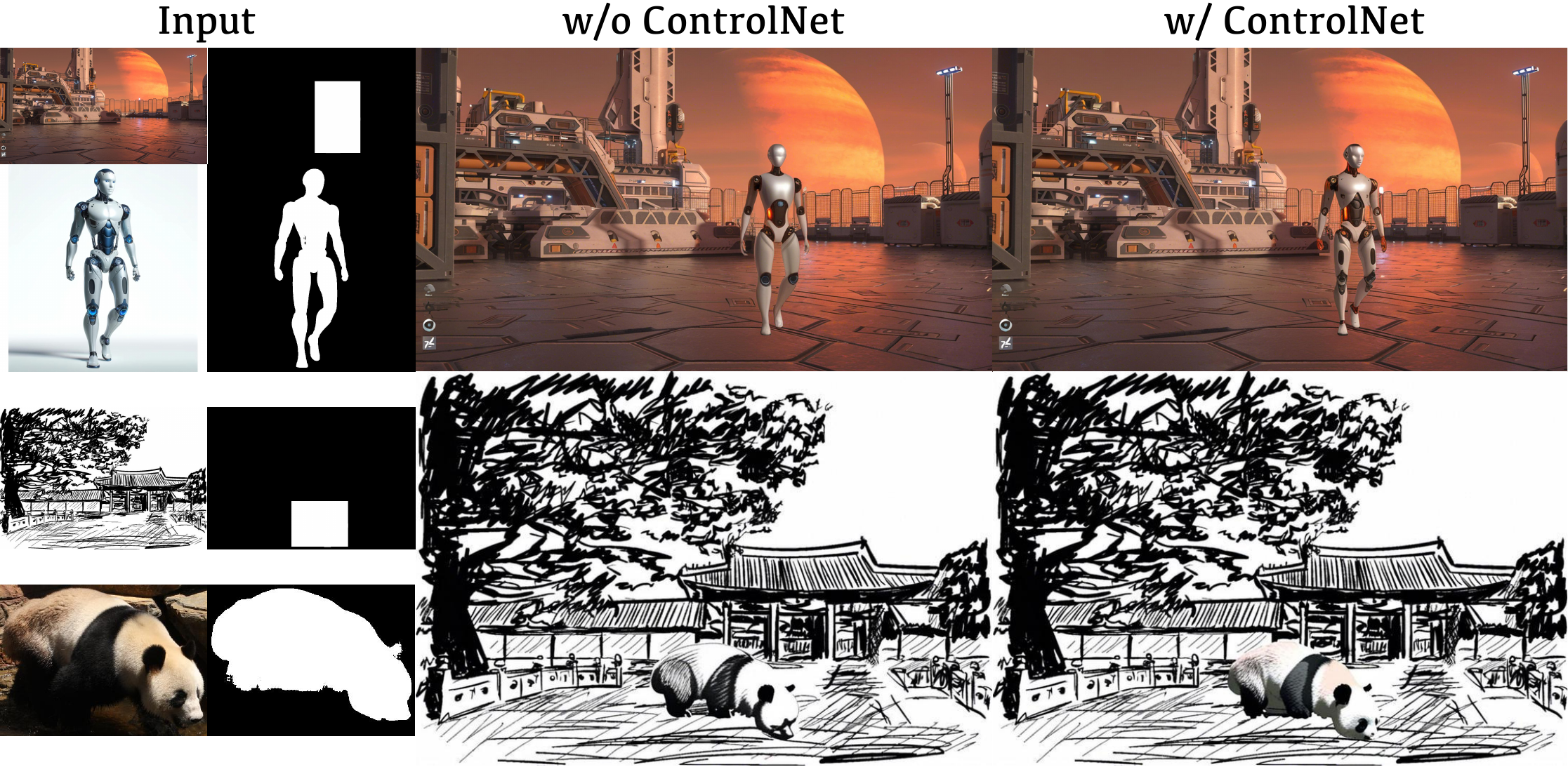}
   \caption{Two examples of our method with and without ControlNet. The top example demonstrates a positive effect on enhancing the details of the robot, while the bottom example shows a negative effect by hindering the stylization.}
   \label{fig:controlnet}
\end{figure}

The proposed method can be readily combined with ControlNet~\cite{controlnet}, since it involves no modification of the main diffusion process. Figure~\ref{fig:controlnet} presents two examples with and without ControlNet.
We use the canny edges~\cite{cannyedge} of the foreground images for additional control. In some cases, such as the top example in Figure~\ref{fig:controlnet}, this conditional strategy enhances the details. However, in other cases, such as the bottom example, rigid control with canny edges may undermine the style coherence of the foreground. Additionally, integrating ControlNet introduces extra computational costs.  

\subsection{Exchanging foreground and background}
\begin{figure}[htbp!]
  \centering
    \includegraphics[width=1\linewidth]{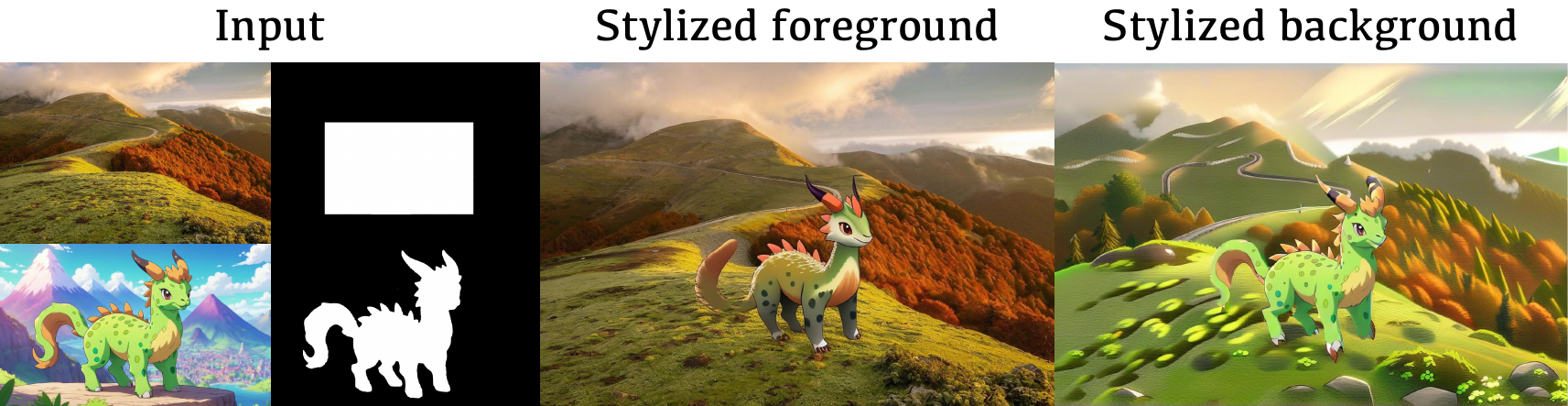}
   \caption{An example of exchanging the roles of the foreground and the background images. Our method allows easy stylization of the foreground with reference to the background, and vice versa.}
   \label{fig:inverse}
\end{figure}

Another distinguishing feature of our method, compared to existing methods, is its \textbf{symmetric} treatment of foreground and background images. This symmetry allows for straightforward generalization to various tasks, such as image stitching where the roles of input images can be easily interchanged. We can also exchange the role of the foreground and background, \ie blending the background into the foreground. Figure~\ref{fig:inverse} (right column) shows an example. We can see that the background is well stylized while the foreground remains exactly unchanged. This task is somewhat different from the image composition task (left) and is useful in practical applications.

\section{More details}

\begin{figure}[htbp!]
  \centering
    \includegraphics[width=1\linewidth]{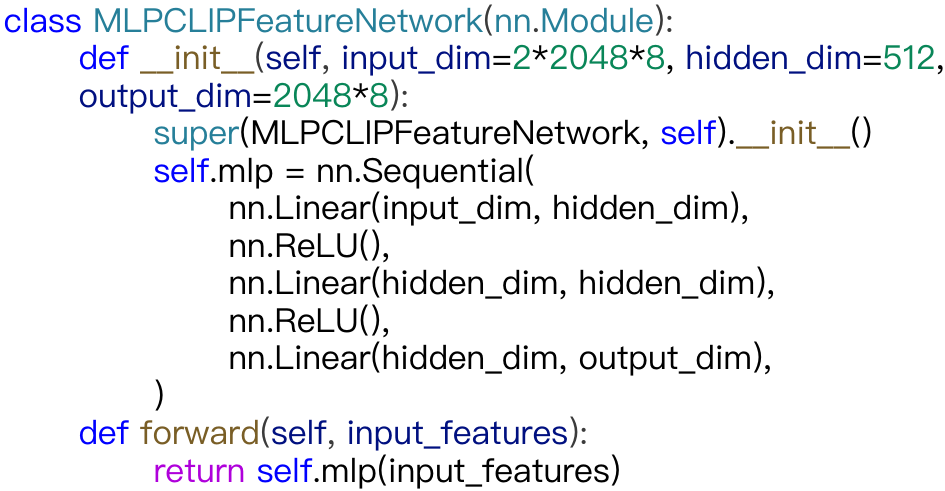}
   \caption{Codes for the $3$-layer MLP network.}
   \label{fig:mlp-code}
\end{figure}

\paragraph{MLP network architecture.} Figure~\ref{fig:mlp-code} shows a simple implementation of the MLP network architecture in PyTorch. 

\paragraph{Principles of Filtering Training Data for MLP.} 

We generate content-style stylized triplets by leveraging a state-of-the-art neural-style transfer method~\cite{csgo}, resulting in a preliminary collection of $65,429$ triplets. The quality and erroneous samples are evaluated on three criteria.
\begin{enumerate}[label=(\alph*)]
    \item \textbf{Content inconsistency}: Removal of triplets where the primary subject in the stylized output deviates significantly from the content image.
    \item \textbf{Style discrepancy}: Exclusion of the stylized results exhibiting mismatches in critical style attributes (e.g., color palette, texture patterns, and brushstroke characteristics) compared to the style image.
    \item \textbf{Substandard visual quality}: Filter artifact-containing outputs with noticeable distortions, blurring, or degradation in perceptual fidelity. 
\end{enumerate}
Finally, we get $37,445$ rigorously filtered style transfer instances.

\paragraph{Questions in user study.}
We randomly shuffle the results from different methods to ensure anonymity in our user study. Given the foreground and background images, as well as their corresponding masks, we ask the users to select the best results referring to the following questions.
\begin{enumerate}[label=(\alph*)]
    \item \textbf{Content preservation}: The synthetic image should retain the characteristics of the foreground subject (such as its identity, shape, and outline) well, without considering color changes.
    \item \textbf{Style consistency}: The style of the foreground subject in the composite image should be consistent with the style of background image, taking into account elements such as tone, texture, and line patterns.
    \item \textbf{Seamless blending}: The composite image needs to seamlessly blend the foreground and background, making it difficult to detect any signs of stitching. There should be no visible edges, artifacts, irregular color patches, or obvious mapping traces.
\end{enumerate}

\paragraph{Computation of CLIP-T.}
In the samples from the TF-ICON~\cite{tf-icon} dataset, where the text prompt describes global image information, we compare the alignment score between the entire resulting image and the given text prompt. However, in most examples from our new benchmark, we compare the alignment score between the locally masked region of the resulting image and the given text prompt. This approach considers that the text prompt, such as 'a dog', describes local information.

